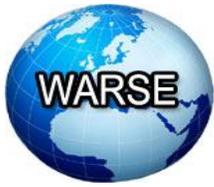

# Empirical Study of Artificial Fish Swarm Algorithm

Reza Azizi
Engineering Department, Bojnourd Branch, Islamic Azad University
Bojnourd, Iran
Email: reza.azizi@bojnourdiau.ac.ir

**ABSTRACT**

Artificial fish swarm algorithm (AFSA) is one of the swarm intelligence optimization algorithms that works based on population and stochastic search. In order to achieve acceptable result, there are many parameters needs to be adjusted in AFSA. Among these parameters, visual and step are very significant in view of the fact that artificial fish basically move based on these parameters. In standard AFSA, these two parameters remain constant until the algorithm termination. Large values of these parameters increase the capability of algorithm in global search, while small values improve the local search ability of the algorithm. In this paper, we empirically study the performance of the AFSA and different approaches to balance between local and global exploration have been tested based on the adaptive modification of visual and step during algorithm execution. The proposed approaches have been evaluated based on the four well-known benchmark functions. Experimental results show considerable positive impact on the performance of AFSA.

## 1. INTRODUCTION

Solving the NP-complete problems is one of the most challenging issues that computer scientists always faced with. Swarm intelligence algorithms have been significantly proved their capabilities in solving these problems. Particle Swarm Optimization (PSO) [1] and Ant Colony Optimization (ACO) [2] are two most well-known methods. These algorithms have some characteristics that make them suitable for solving NP-complete problems, like scalability, fault tolerance, consistency, higher speed, flexibility and parallelism.

Artificial fish swarm algorithm (AFSA) [3], proposed by Li Xiao Lei in 2002, is a stochastic population-based algorithm motivated by intelligent collective behavior of fish groups in nature. AFSA has characteristics of non-sensitive initial artificial fish location, flexibility and fault tolerance. It has been applied on different problems including machine learning [4,5,6], PID controlling [7], wireless sensor networks [8], image segmentation [9], data clustering [10,11] and scheduling [12].

AFSA did not gain much acceptance due to certain due to certain reasons. Among them high computational complexity, difficult implementation of the algorithm and the results not significantly better than similar algorithms can be noted here. In fact, algorithms such as PSO with less computational complexity are easier to implement and exploit. Furthermore, results obtained from different versions of PSO show better performances compare to standard AFSA.

Contrary to what may appear, AFSA is not a version of PSO and differs significantly from PSO. One of the main differences between two algorithms is that particles in PSO move completely based on their past movements and experiences in the problem environment. However, artificial fish movements solely depend on their current positions and other members of the group situations. Hence, movements of the fish differ from particles.

There are two important parameters in AFSA: visual and step. Artificial fish search the problem environment as large as their visual. Afterwards, they move towards the target based on the random value of the step in each iteration. In standard AFSA, initial values for these parameters have a great effect on the final result, because of the fact that these parameters remain constant and equal to the initial values until the algorithm terminates. If we select larger initial values for visual and step, artificial fish swarm move faster toward the global optimum and will be more capable of passing the local optimums. And selecting lower values for these parameters causes artificial fish to act better in local searching.

In this paper, improved artificial fish swarm algorithm has been proposed with a new parameter, called movement weight. Movement weight can be used to adjust visual and step adaptively and consequently controls the movements of artificial fish towards the target. Moreover, using the movement weight considerably maintains equilibrium between global and local searches.

Beforehand, parameter called inertia weight has been applied on particle swarm optimization (PSO) algorithm by Shi and Eberhart [13]. Inertia weight, which can be a positive constant number [13], or a positive linear or non-linear function [14], or others [15], significantly improved the performance of algorithm. Our proposed movement weight in AFSA plays a similar role as inertia weight in PSO. In this paper, different versions of movement weight have been investigated, including a positive constant weight, random and positive linear functions. Evaluating the applied adaptive movement weight on four well known benchmark functions shows considerable improvement in performance of the AFSA.

Subsequent parts of this paper have been organized as follows: section 2 briefly introduces the standard AFSA, afterward, AFSA with variable visual and step will be described in section 3. Section 4 studies the experimental results and finally we conclude our work in the last section.

## 2. ARTIFICIAL FISH SWARM ALGORITHM

In underwater world, fish can find areas with more food based on their individual or swarm search. Inspired by this characteristic, Artificial Fish (AF) model is represented by prey, free move, swarm, and follow behaviors. AF searches the problem space by those behaviors. The environment AF lives in, is the problem space. Objective function of AFSA is to find maximum food density.

As Figure 1 shows, AF observes external concepts with its visual perception. Current position of AF is shown by vector $X=(X_1, X_2, …, X_n)$. The visual is equal to visibility domain of AF and $X_v$ is an intended position in visual where the AF selects to move towards. If $X_v$ has better food density than current position, AF moves one step towards $X_v$, which results in displacement of AF from X to $X_{next}$. Otherwise, if the current position of AF is better than $X_v$, it selects another position in its visual. Food density in position X is the fitness value of the position and is shown with $f(X)$. The step is equal to maximum length of the movement. The distance between two AFs which are placed in $X_i$ and $X_j$ is shown by (Euclidean distance) $Dis_{(ij)}=\| X_i - X_j \|$.

AF model consists of variables and functions. Variables are X (current AF position), *step* (maximum length step), *visual* (visibility domain), *try-number* (maximum attempts for finding better positions in visual), and crowd factor δ (0<δ<1). Functions consist of *prey*, *free move*, *swarm*, and *follow* behaviors.

In each iteration of the optimization process, AF looks for locations with better fitness values in problem space by performing one of these behaviors. We omit to explain details of artificial fish due to space limitation. Readers are referred to [3, 6, 7] for more information.

## 3. ARTIFICIAL FISH SWARM ALGORITHM WITH VARIABLE VISUAL AND STEP

In AFSA, artificial fish search the problem environment in their visual based on the behaviors and then they move towards the target by a random value of their step. In standard AFSA, determination of the initial values of the step and visual essentially influence the final result. Values of these parameters remain constant and equal to the initial values during the algorithm execution. By having large initial values for these parameters, artificial fish swarm move faster towards the global optimum. That is, artificial fish are able to search larger environment around them and move with bigger step. Therefore, artificial fish are more capable in escaping from the local optimums under such circumstances. However, there are some deficiencies in large values of step and visual. Accuracy and steadiness of the algorithm decrease in such situation.

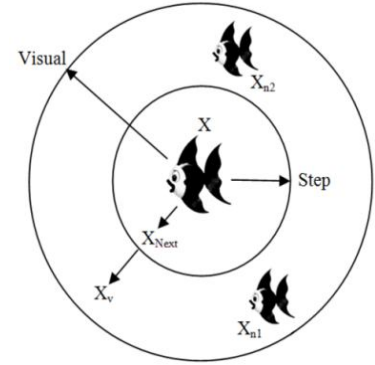

**Figure 1:** Artificial Fish and the Environment.

In fact, the algorithm acts better in global searches, but after approaching the global optimum, it is incapable of doing an appropriate local search because of the fact that the visual is larger than it must be. Therefore, by large value of the visual, positions with better fitness are unlikely to be found. Considering small values for these parameters make algorithm more steady and accurate, nevertheless, it causes the algorithm move towards the target more slowly and be incapable of escaping local optimums.

Based on the above facts, in order to attain better results, larger initial values for visual and step must be selected and then gradually they must be adaptively reduced during the algorithm execution. As a result, fish move towards to the target quickly and are more capable of escaping local optimums. Afterwards, upon approaching global optimum, artificial fish can accurately investigate the environment by smaller visual and step.

In order to control values of step and visual and balancing between global search and local search, we propose a parameter called Movement Weight (MW). MW can be a constant value smaller than one, positive linear or nonlinear function. In each iteration, visual and step values are given according to the following equations:

$$Visual_{Itr} = MW \times Visual_{Itr-1} \qquad (1)$$

$$Step_{Itr} = MW \times Step_{Itr-1} \qquad (2)$$

Where *Itr* is the current iteration of the algorithm. With the purpose of attaining better values for the visual and step based on the iteration number, different methods for calculating MW have been discussed in this study.

## 3.1 Linear Movement Weight

In this method, MW is a positive function varies between a minimum and a maximum and calculated according to current iteration and final iteration number. In linear-decreasing MW [16], at the beginning of the algorithm, MW is equal to the maximum value. During the algorithm execution this amount linearly reduces to the minimum value based on the iteration number. MW in each iteration is calculated according to the Equation 3.

In the linear-increasing MW [17], fist MW is set to the minimum value, then it increases linearly during the execution and finally it reaches maximum value. MW in the each iteration is given according to the Equation 4:

$$MW_{Itr} = MW_{Min} + \frac{Itr_{Max} - Itr}{Itr_{Max}} \times (MW_{Max} - MW_{Min}) \quad (3)$$

$$MW_{Itr} = MW_{Max} - \frac{Itr_{Max} - Itr}{Itr_{Max}} \times (MW_{Max} - MW_{Min}) \quad (4)$$

where *Rand* is a random value between 0 and 1 with normal distribution.

## 3.2 Random Movement Weight

In random MW [18], MW is a random number between two values: minimum and maximum. In each iteration, Random value of MW is calculated based on the following equation:

$$MW_{Itr} = MW_{Min} + Rand \times (MW_{Max} - MW_{Min}) \quad (5)$$

## 4. EXPERIMENTS AND ANALYSIS

Experiments are performed on four benchmark functions that are often used as measurement criteria of optimization algorithms in continuous and static spaces. Benchmark functions with their search space range and acceptable solution are presented in Table 1 [15]. It should be noted that optimal value of all these function equals zero.

Initial values for visual and step has been considered 40% and 25% of the range length of the fitness function variables respectively (For example range length of the fitness function variables in Ackly function is 64). Crowd factor is 0.5, maximum try-number is 10 and population number is 30.

Determination of the MW plays a vital role in accuracy and quality of the final results. Small value of the MW may cause a sharp reduction in the amount of visual and step, which causes the artificial fish become stationary and blind before reaching the global optimum.

In fact, artificial fish cannot effectively search the environment when the amount of visual is too small. This is also true about the step. Actually, when the artificial fish are not close to the global optimum and step value is too small, artificial fish cannot efficiently move inside the problem environment. Consequently, selecting a small MW value makes artificial fish stationary.

Determination of MW depends on the type of the problem and the number of iterations. Table 2 shows the optimal values of the MW for 100, 200, 500, 1000, 1500 and 2000 iterations for the mentioned benchmark functions in 30-dimensional space. In all cases different constant values for the MW have been assumed and the best MW value for the different number of iterations has been obtained according to the results. Experiments for each iteration have been done on different values of MW = (0.72, 0.73, 0.74 ... 0.9, 0.91 ... 1.00, 1.01, 1.02). Algorithm has been run 50 times for the each iteration number, and average results for the Sphere, Ackley, Rosenbrock and Griewank functions have been shown in the Table 2.

**Table 1:** Attributes of Cleveland dataset.

| Name | Function | Acceptance | Search space |
|---|---|---|---|
| Sphere | $f_1(x) = \sum_{i=1}^{D} x_i^2$ | 0.01 | [-500,500] |
| Rosenbrock | $f_2(x) = \sum_{i=1}^{D} \left(100(x_{i+1} - x_i^2)^2 + (x_i - 1)^2\right)$ | 100 | [-10,10] |
| Ackly | $f_3(x) = 20 + e - 20e^{-0.2\sqrt{\frac{1}{n}\sum_{i=1}^{n} x_i^2}} - e^{\frac{1}{n}\sum_{i=1}^{n} \cos(2\pi x_i)}$ | 0.01 | [-32,32] |
| Griewank | $f_4(x) = \sum_{i=1}^{n} \left(\frac{x_i^2}{4000}\right) - \prod_{i=1}^{n} \cos\left(\frac{x_i}{\sqrt{i}}\right) + 1$ | 0.01 | [-600,600] |

**Table 2.** Best MW and corresponding mean results in 50 times execution for different iteration numbers for Sphere, Rosenbrock, Ackly, and Griewank functions.

| Functions | | 100 Itr | 200 Itr | 500 Itr | 1000 Itr | 1500 Itr | 2000 Itr |
|---|---|---|---|---|---|---|---|
| $F_1$ | MW | 0.94 | 0.95 | 0.96 | 0.96 | 0.96 | 0.96 |
| | Result | 0.7633 | 1.122e-04 | 7.939e-14 | 1.523e-31 | 2.831e-49 | 5.208e-67 |
| $F_2$ | MW | 0.95 | 0.96 | 0.96 | 0.96 | 0.96 | 0.97 |
| | Result | 57.2397 | 46.8643 | 41.8349 | 37.0577 | 27.7131 | 27.6717 |
| $F_3$ | MW | 0.95 | 0.96 | 0.96 | 0.96 | 0.97 | 0.98 |
| | Result | 1.2416 | 0.0154 | 9.376e-06 | 1.521e-14 | 6.572e-15 | 3.908e-15 |
| $F_4$ | MW | 0.94 | 0.96 | 0.97 | 0.98 | 0.98 | 0.98 |
| | Result | 0.1165 | 0.0013 | 1.850e-10 | 1.213e-15 | 1.036e-16 | 0 |

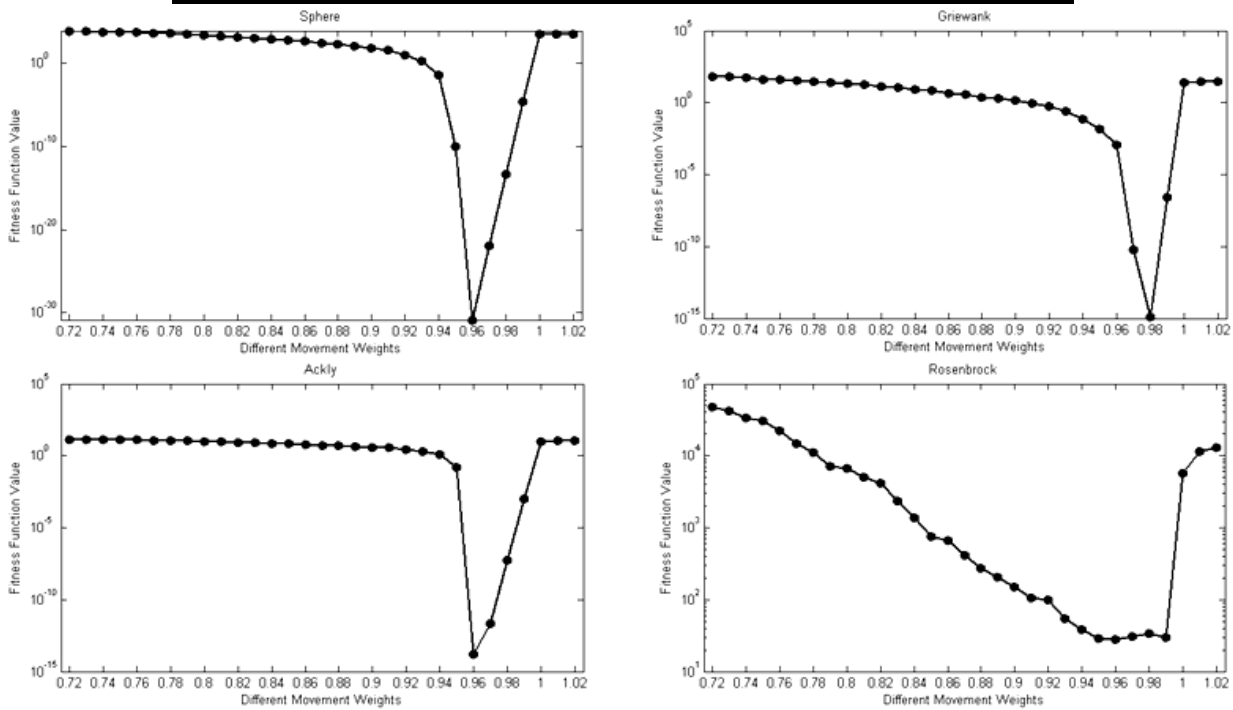

**Figure 2.** The average results obtained on different MWs of 30-Dimentional 4 benchmark function in 1000 iterations.

Figure 2 shows the average results of 50 times of algorithm execution for the best value of MW for each benchmark function in 1000 iterations.

The best MW range and the best MW that generate acceptable results for Sphere, Ackly, Rosenbrock and Griewank for 1000 iterations have been depicted in Table 3. Experiments repeated 50 times and mean, best and standard deviation obtained from running standard AFSA, Constant Weight AFSA (CWAFSA), Random Weight AFSA (RWAFSA), Linear-Decreasing Weight AFSA (LDWAFSA), Linear-Increasing Weight AFSA (LIWAFSA) and global version of PSO (GPSO) [13] in 10, 20 and 30-dimensional spaces on Sphere, Ackly, Rosenbrock and Griewank have been summarized on Table 4. The lower bound and the upper bound values of the MW for RWAFSA, LIWAFSA and LDWAFS for each function have been selected based on the Acceptable Movement Weights Range column of the Table 3. MW value for CWAFSA for each function has been attained based on the Best Movement Weight column of Table 3 as well. Inertia weight value, in GPSO, linearly decreases from 0.9 to 0.4 during the algorithm execution [15]. The population size is equal to 5 * D where D is the problem dimension. At last, parameters C1 and C2 have been set in the form of $C_1 = C_2 = 2$.

As it can be seen from Table 4 and Figure 3, MW parameter can considerably improve the results in AFSA. Among different proposed algorithms in this paper, LIWAFSA reaches acceptable results, but with less performance than other algorithms. In LIWAFSA, values of visual and step parameters experience a sharp fall at the beginning of the algorithm, which makes artificial fish unable to search and move properly after a short period of time. Determination of the proper MW value in CWAFSA helps the algorithm reaching very acceptable results. Consequently in general, among the proposed algorithms, CWAFSA attain reasonable results if the MW has been chosen appropriately.

LDWAFSA algorithm shows slow progress at the beginning of the algorithm execution, because at the beginning of the algorithm the value of MW remains near the maximum amount of weight and Visual and Step parameters witness a smooth decrease in value.

After a while, MW becomes smaller and the visual and step values decrease with a higher rate rather than previous iterations which leads in faster progress of the algorithm. In RWAFSA, the MW value is determined randomly which increase in diversity of the group movement and leads to uniform progress of the algorithm. On the whole, in different applications we can attain better results with defining the optimal weight for CWAFSA if it is possible.

**Table 3.** Best and acceptable range of MW for 1000 iteration.

| Function | Best Movement Weight | Acceptable Movement Weights Range |
|---|---|---|
| Sphere | 0.96 | [0.95 , 0.99] |
| Rosenbrock | 0.96 | [0.93 , 0.99] |
| Ackly | 0.96 | [0.95 , 0.99] |
| Griewank | 0.98 | [0.94 , 0.99] |

**Table 4.** Comparison of best, mean and standard deviation results of Standard AFSA, GPSO, CWAFSA, LDWAFSA, LIWAFSA and RWAFSA for 50 times execution for Sphere, Rosenbrock, Ackly and Griewank functions in 10, 20 and 30 dimensional space for 1000 iterations.

| F | D | Criteria | StdAFSA | PSO | CWAFSA | RWAFSA | LDWAFSA | LIWAFSA |
|---|---|---|---|---|---|---|---|---|
| Sphere | 10 | Best | 0.138 | 8.215e-29 | **5.487e-35** | 4.636e-26 | 7.647e-26 | 3.180e-26 |
| | | Mean | 0.330 | 2.397e-25 | **1.772e-34** | 2.500e-25 | 3.033e-25 | 1.192e-25 |
| | | Std-dev | 0.163 | 9.991e-25 | **6.859e-35** | 2.103e-25 | 1.340e-25 | 5.232e-26 |
| | 20 | Best | 2.724 | 1.062e-13 | **3.833e-33** | 1.942e-24 | 8.069e-24 | 7.598e-25 |
| | | Mean | 5.567 | 1.925e-11 | **8.155e-33** | 1.022e-23 | 1.380e-23 | 2.225e-24 |
| | | Std-dev | 2.185 | 2.800e-11 | **2.211e-33** | 1.291e-23 | 3.575e-24 | 9.342e-25 |
| | 30 | Best | 378.428 | 1.026e-07 | **3.048e-32** | 4.327e-24 | 4.777e-23 | 1.033e-23 |
| | | Mean | 666.478 | 6.287e-06 | **3.997e-32** | 3.226e-23 | 9.124e-23 | 1.734e-23 |
| | | Std-dev | 153.831 | 6.468e-06 | **4.788e-33** | 2.170e-23 | 2.329e-23 | 6.721e-24 |
| Rosenbrock | 10 | Best | 7.289 | 1.047 | 0.208 | **0.075** | 1.196 | 1.543 |
| | | Mean | 43.294 | 1.082e+02 | **5.190** | 16.487 | 12.212 | 8.693 |
| | | Std-dev | 70.258 | 2.297e+02 | **1.708** | 60.272 | 21.018 | 12.773 |
| | 20 | Best | 72.746 | 16.086 | **14.122** | 15.185 | 15.766 | 15.946 |
| | | Mean | 262.724 | 1.846e+03 | **32.494** | 35.024 | 34.006 | 54.972 |
| | | Std-dev | 153.927 | 4.839e+03 | **30.577** | 36.376 | 44.314 | 65.069 |
| | 30 | Best | 5.099e+4 | 44.962 | **24.570** | 25.931 | 26.374 | 82.986 |
| | | Mean | 1.639e+5 | 2.142e+04 | 59.776 | **54.153** | 80.257 | 249.777 |
| | | Std-dev | 8.696e+4 | 3.342e+04 | **44.452** | 45.440 | 143.095 | 91.093 |
| Ackly | 10 | Best | 0.114 | 9.769e-15 | **2.664e-15** | 7.371e-14 | 1.021e-13 | 1.092e-13 |
| | | Mean | 0.980 | 3.677e-13 | **2.664e-15** | 3.154e-13 | 3.327e-13 | 2.150e-13 |
| | | Std-dev | 0.748 | 9.615e-13 | **0** | 2.032e-13 | 2.085e-13 | 6.196e-14 |
| | 20 | Best | 1.546 | 8.961e-08 | **2.664e-15** | 3.508e-13 | 8.482e-13 | 2.264e-13 |
| | | Mean | 2.690 | 1.258e-06 | **2.901e-15** | 7.021e-13 | 1.060e-12 | 4.337e-13 |
| | | Std-dev | 0.568 | 9.412e-07 | **9.013e-16** | 2.962e-13 | 1.312e-13 | 1.061e-13 |
| | 30 | Best | 5.637 | 1.120e-04 | **1.332e-14** | 7.665e-13 | 1.697e-12 | 7.878e-13 |
| | | Mean | 6.814 | 8.418e-04 | **1.592e-14** | 1.432e-12 | 2.222e-12 | 0.005 |
| | | Std-dev | 0.519 | 0.001 | **5.895e-15** | 5.123e-13 | 1.968e-13 | 0.020 |
| Griewank | 10 | Best | 0.009 | 6.864e-11 | **0** | **0** | **0** | **0** |
| | | Mean | 0.043 | 0.003 | 3.700e-17 | 6.661e-17 | **1.924e-16** | 5.460e-09 |
| | | Std-dev | 0.031 | 0.017 | 8.416e-17 | **1.034e-16** | 2.590e-16 | 2.257e-08 |
| | 20 | Best | 0.072 | 3.258e-07 | **0** | **0** | 4.440e-16 | 7.951e-07 |
| | | Mean | 0.192 | 0.037 | **1.702e-16** | 3.237e-14 | 1.162e-15 | 6.094e-04 |
| | | Std-dev | 0.078 | 0.077 | **1.119e-16** | 1.209e-13 | 5.237e-16 | 9.454e-04 |
| | 30 | Best | 5.159 | 1.122e-05 | **2.220e-16** | 1.332e-15 | 1.643e-13 | 0.004 |
| | | Mean | 7.134 | 0.070 | **4.292e-16** | 2.244e-06 | 3.453e-10 | 0.036 |
| | | Std-dev | 1.264 | 0.125 | **2.486e-16** | 6.605e-06 | 1.224e-09 | 0.032 |

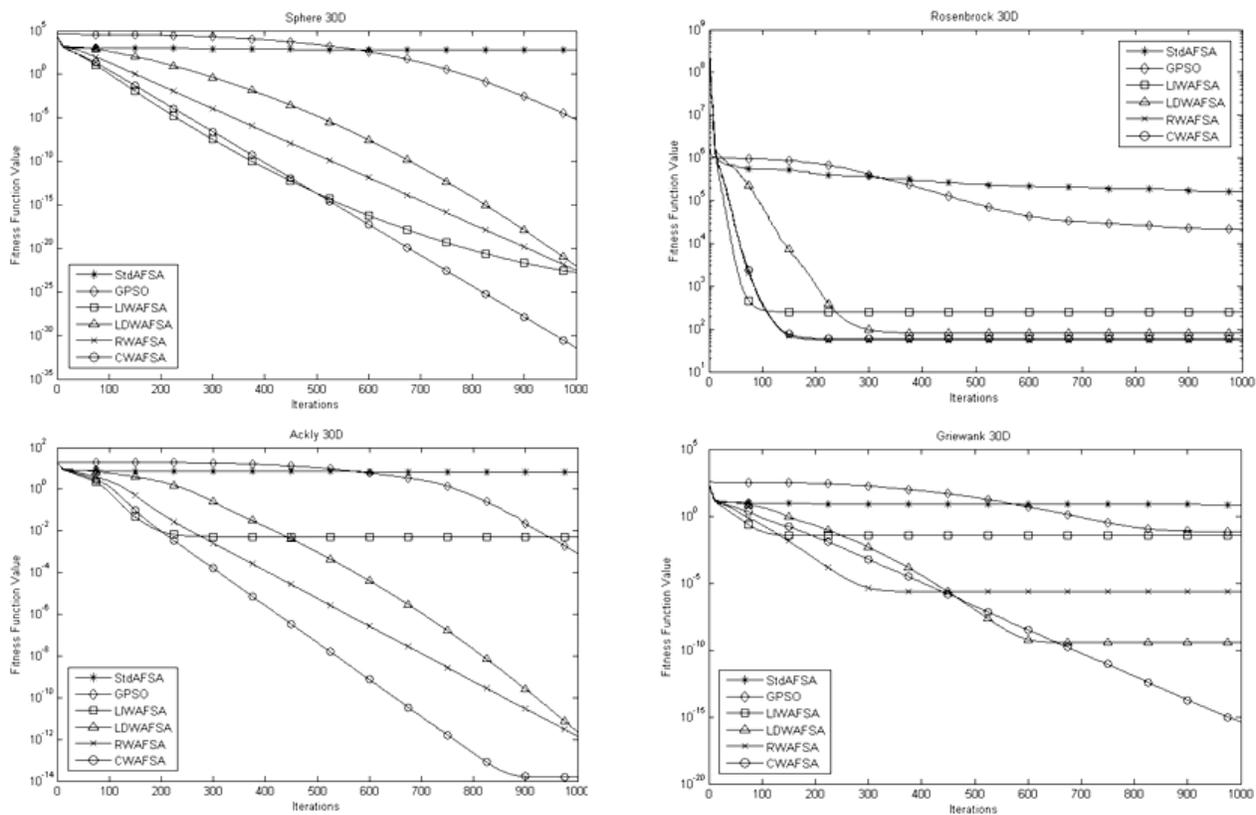

**Figure 3.** Comparison of the convergence behavior of 6 algorithms on 4 tested benchmark functions in 30D space.

But if it is impossible to determine optimal weight, one can use RWAFSA and LDWAFSA by determination of a suitable range of MW values. Finally according to the results, AFSA equipped with MW generates results analogous to PSO. As results show, in higher dimensions, modified AFSA acts even better than PSO with inertia weight.

## 5. CONCLUSION

In this paper, we have applied a parameter called Movement Weight (MW) on the Artificial Fish Swarm Algorithm (AFSA). Experimental results reveal the positive impact of this parameter on the performance of AFSA. It can be easily seen that AFSA with an appropriate MW parameter can act better than standard AFSA. Moreover, time varying MW and a random MW result a significant improvement on the AFSA performance. Various simulations have been done to support the competence of the MW.